\documentclass[conference]{IEEEtran}
\IEEEoverridecommandlockouts
\usepackage{amsmath,amssymb,amsfonts}
\usepackage{algorithmic}
\usepackage{graphicx}
\usepackage{textcomp}
\usepackage{xcolor}
\usepackage[utf8]{inputenc}
\usepackage[vietnamese=nohyphenation]{hyphsubst}
\usepackage[vietnamese,english]{babel}
\usepackage[sorting=none]{biblatex}
\usepackage{multirow}
\usepackage{csquotes}
\def\BibTeX{{\rm B\kern-.05em{\sc i\kern-.025em b}\kern-.08em T\kern-.1667em\lower.7ex\hbox{E}\kern-.125emX}}
\begin{document}

\title{Exploiting Vietnamese Social Media Characteristics for Textual Emotion Recognition in Vietnamese}


\author{\textbf{Khang Phuoc-Quy Nguyen}$^{1,2}$ and \textbf{Kiet Van Nguyen}$^{1,2}$\\$^{1,2}$University of Information Technology, Ho Chi Minh City, Vietnam\\$^{1,2}$Vietnam National University, Ho Chi Minh City, Vietnam\\{16520568@gm.uit.edu.vn, kietnv@uit.edu.vn}}

\maketitle

\begin{abstract}
Textual emotion recognition has been a promising research topic in recent years. Many researchers aim to build more accurate and robust emotion detection systems. In this paper, we conduct several experiments to indicate how data pre-processing affects a machine learning method on textual emotion recognition. These experiments are performed on the Vietnamese Social Media Emotion Corpus (UIT-VSMEC) as the benchmark dataset. We explore Vietnamese social media characteristics to propose different pre-processing techniques, and key-clause extraction with emotional context to improve the machine performance on UIT-VSMEC. Our experimental evaluation shows that with appropriate pre-processing techniques based on Vietnamese social media characteristics, Multinomial Logistic Regression (MLR) achieves the best F1-score of 64.40\%, a significant improvement of 4.66\% over the CNN model built by the authors of UIT-VSMEC (59.74\%).
\end{abstract}

\begin{IEEEkeywords}
Textual Emotion Recognition, Emotion Prediction, Vietnamese Characteristics, Machine Learning, Multinomial Logistic Regression.
\end{IEEEkeywords}

\section{Introduction}\label{Introduciton}
Analyzing people’s opinions has never been an easy task due to their versatility and ambiguity. Answering the question “What do people think about X?” is quite simple; we achieve this by extracting information from what they said or wrote, but to understand their feelings or answer the question “How do people feel about X?” is a difficult task. There are many ways to express the same emotion, but multiple emotions sometimes have the same expression. For example, a smile does not always mean happy, but it can be an act of disgust; people may cry when they feel sad, or when they are afraid of something or even when they are happy. When it comes to emotion recognition using a computer, this problem is more complicated. Emotion detection can be performed based on different sources of information such as speech, text, or image, but text data is still the most popular form, especially in social media. Textual emotion recognition has seen a growing interest from many researchers \cite{brilis2012mood,ho2012high,ghazi2014prior,da2014tweet,ho2019emotion}.

There are several approaches to build an emotion detection system, but they all have to go through one step: data pre-processing. Data pre-processing is the process that aims to remove as much as possible noise from the data. Therefore, a variety of pre-processing techniques have been tested by many researchers, and they are proven to be efficient for some particular machine learning tasks.

In this work, we will focus on evaluating the impact of several pre-processing techniques on the emotion recognition system’s performance built on Vietnamese social media data (Vietnamese microtext \cite{ellen2011all}). With a given Vietnamese microtext, our goal is to clean this microtext and then predict the emotion it contains. 

Our key contribution is to compile an appropriate set of pre-processing techniques for Vietnamese social media data. We build several classifiers for a different combination of pre-processing techniques by using Multinomial Logistic Regression (MLR) \cite{bohning1992multinomial} on the benchmark dataset UIT-VSMEC \cite{ho2019emotion}. Finally, we compare the F1-score of each classifier to find out the best combination of pre-processing techniques.

The structure of this paper is as follows. Related studies are presented in Section \ref{Related-work}. Section \ref{Dataset} describes the dataset used in our experiments. Section \ref{Methodology} discusses our proposed methods in which we explain the reasons behind each method. In Section \ref{Experiments}, we present the experiments and results for our proposed method. Finally, we sum up our findings and discuss future work in Section \ref{conclusion}.

\section{Related Work}\label{Related-work}

In the past decade, there were many studies on the design of methods and the development of frameworks for textual emotion detection. A detailed and complete overview of approaches can be found in \cite{sailunaz2018emotion,calvo2010affect}.

\textcite{brilis2012mood} used machine learning approaches to create an automatic mood categories classifier for song lyrics by performing pre-processing steps such as stemming, stop words, and punctuation marks removal combined with a bag-of-words approach using TF-IDF score (term frequency-inverse document frequency). The authors indicated that the Random Forest algorithm reported the best results, with approximately 71.50\% accuracy on the stemmed dataset and 93.70\% on the unstemmed.

\textcite{ho2012high} proposed a hidden Markov logic approach to predict the emotion expressed by a given text. The authors’ method achieved an F-score of 35.00\% on the ISEAR (International Survey on Emotion Antecedents and Reaction) dataset. The authors ignored semantic and syntactic features of the sentence analysis, which made it non-context sensitive and can be the cause for the low F-score.

\textcite{ghazi2014prior} combined keyword-based, lexicon-based, and machine learning-based approaches to compare the prior emotion of a word with its contextual emotion and their effect on the sentiment expressed by the sentence. Authors showed that using Logistic Regression and their features significantly outperform their baselines (the null hypothesis that was using the least and the most frequent emotion label to predict) and their Support Vector Machines (SVM) model.

\textcite{da2014tweet} conducted tweet sentiment analysis using classifier ensembles. A classifier ensemble was formed using several machine learning classifiers: Random Forest, SVM, Multinomial Naive Bayes, and Logistic Regression. In their study, authors experimented with various tweet datasets and reported that the classifier ensemble could improve classification accuracy. 

In work presented in \cite{ho2019emotion}, the authors built the first standard Vietnamese Social Media Emotion Corpus (UIT-VSMEC). This corpus contains exactly 6,927 emotion-annotated sentences (an average agreement of 82.00\% across the seven emotions) and then measure two machine learning models and two deep learning models on this dataset. The authors reported the highest weighted F1-score of 59.74\% by using the CNN model with pre-trained word2Vec.

Although many efforts had been made in building machine learning models, pre-processing has not been adequately addressed for the Vietnamese language yet, and little research was focused on this. Therefore, we decide to experiment on how different pre-processing techniques contribute to the emotion classifier’s performance. In this paper, we will evaluate MLR models on UIT-VSMEC.

\section{Dataset}\label{Dataset}
In this paper, we conduct our experiments on UIT-VSMEC. This dataset contains 6,927 Facebook comments with each comment only assigned an emotion label. There are seven emotion labels in this dataset consisting of Ekman’s six basic emotions labels \cite{ekman1999basic} and “Other” label if a comment does not include any emotion or if a comment contains emotion that is different from Ekman’s six basic emotions labels. We follow the training, development and test sets of the dataset for conducting our experiments.




\section{Methodology}\label{Methodology}
Our task belongs to textual emotion recognition, which aims to identify human emotion from the text. We separate our task into a pre-processing problem, which prepares the data for our training purpose, and a key-clauses extraction problem, which detects the main clause that is most likely to carry emotion in the given comment.

\subsection{Pre-processing}\label{pre-processing}
Since the author of UIT-VSMEC had not performed much cleaning on the data, we come up with several pre-processing techniques for Vietnamese social media data (Vietnamese microtext) \cite{satapathy2017phonetic}. Table \ref{tab2} describes seven pre-processing techniques based on Vietnamese social media characteristics.

\begin{table}[htbp]
\centering
\caption{Pre-processing techniques.}
\label{tab2}
\begin{tabular}{cl}
\hline
{\bf Technique} & \multicolumn{1}{c}{\bf Description}                                                                   \\ \hline
1         & Standardizing words.                                                                                  \\ \hline 
2         & Removing emojis, emoticons.                                                                            \\ \hline
3 & \begin{tabular}[c]{@{}l@{}}Keeping emojis and emoticons then transforming\\ them into their word form.\end{tabular}                                    \\\hline 
4 & \begin{tabular}[c]{@{}l@{}}Removing repeated emojis and emoticons in the \\ same comment then transforming them into their \\ word form.\end{tabular} \\\hline 
5         & \begin{tabular}[c]{@{}l@{}}Translating word form of emojis, emoticons into \\ Vietnamese.\end{tabular} \\\hline 
6         & \begin{tabular}[c]{@{}l@{}}Spelling correction and acronyms, abbreviations \\ lookup.\end{tabular}  \\\hline 
7         & Removing stop words and particular words.                                                             \\ \hline
\end{tabular}
\end{table}

UIT-VSMEC corpus contains several words that are not in proper formats. For example, “luônnn” should be “luôn” (always) and “thích quáaaaa” should be “thích quá” (really enjoyable).

We partly solve this problem by using technique 1, in which we use regular expressions to replace a sequence of repeated characters by the character itself. For example, “hahaaaa” is converted to “haha” and “:]]]]” is converted to “:]”. We cannot solve this problem entirely because of accents. For example, “thích quáaaaa” should be “thích quá” (really enjoyable), but our method only changes it to “thích quáa”. Table \ref{tab3} shows some examples of the word standardized process. 

\begin{table}[htbp]
\centering
\caption{Example of pre-processing technique 1.}
\label{tab3}
\begin{tabular}{ll}
\hline
{\bf Words} & {\bf Standardized words} \\ \hline
:))))              & :)            \\ 
:{]}{]}{]}         & :{]}          \\ 
“hahaa”            & “haha”        \\ 
“hicc”             & "hic"         \\ 
"luônnn"           & "luôn" (always)    \\ \hline
\end{tabular}
\end{table}

Some many emojis and emoticons appear in our corpus. An emoticon is a pictorial representation of a facial expression using characters - usually punctuation marks, numbers, and letters - to express a person’s feelings or mood. Emojis are much like emoticons, but emojis are pictures rather than typographic approximations. Because emojis are considered as one-letter words and emoticons are considered as a combination of punctuation marks, they will usually be filtered out in the training process despite being the characteristics of social media data.

To handle emojis and emoticons, we come up with three approaches: remove, transform, and hybrid. The first approach is to use technique 2 to remove all the emojis and emoticons in our corpus. Although this technique helps reduce the vocabulary size, it has one big drawback. It considers emojis and emoticons do not express any feeling of the user. Still, most social media users tend to use emojis or emoticons to share their joy, sorrow, or anger.

The second approach is to transform emojis and emoticons into their word form. To convert emoticons into word form, we prepare an emoticon dictionary by changing all emoticons extracted from our corpus. For example, ":)" is replaced by ":slightly\_smiling\_face:" whereas ":(" is replaced by ":frowning\_face:". To transform emojis into word form, we use the demojize function in Python’s emoji package. These are the core of technique 3. This approach considers all emojis and emoticons are important and should not be removed. Table \ref{tab4} shows some examples of the word form of some emojis, emoticons.

\begin{table}[htbp]
\centering
\caption{Several examples for converting emojis and emoticons to its word form.}
\label{tab4}
\centerline{\includegraphics[scale=0.72]{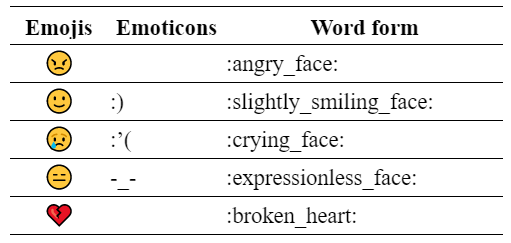}}
\end{table}

The third approach is to remove repeated emojis and emoticons in the same corpus, transforming them into their word form. Technique 4 considers duplicated emojis and emoticons in the same corpus are noise data, and only the unique ones should be kept.

\begin{otherlanguage*}{vietnamese}
The word forms of emojis and emoticons are in English, but the language of our corpus is Vietnamese. This behaviour makes emojis and emoticons have a different meaning with emotional words in the corpus. We use technique 5 to translate word forms of emojis, emoticons into Vietnamese. We create a translation rule dictionary for this purpose. The dictionary contains 109 translation rules for each word form of emojis and emoticons. For example, “disappointed\_face” is translated to “thất vọng” (disappointed). We present some examples in Table \ref{tab5}.

\begin{table}[htbp]
\centering
\caption{Example demojized form of emojis and their Vietnamese form in the translation rule dictionary.}
\label{tab5}
\begin{tabular}{lll}
\hline
{\bf Emotion}             & {\bf Vietnamese} & {\bf English} \\ \hline
:angry\_face:             & tức giận      & angry            \\ 
:slightly\_smiling\_face: & cười nhẹ      & slightly smiling \\ 
:crying\_face:            & khóc          & crying           \\ 
:expressionless\_face:    & không cảm xúc & expressionless   \\ 
:broken\_heart:           & đau lòng      & broken heart     \\ \hline
\end{tabular}
\end{table}

There are a large number of misspelling words, slang words, and abbreviates in UIT-VSMEC. This behaviour is the nature of microtext, as explained in \cite{ellen2011all}. For example, “bjo” should be “bây giờ” (now) and “ju jin” should be “giữ gìn” (conserve).

We correct these spelling errors by compiling an acronym and misspelling dictionary. The compiled dictionary has translations for 736 acronyms and misspelling words. Using the dictionary, we change acronyms and misspelt words to its proper form. Table \ref{tab6} shows some examples of this technique.
\begin{table}[htbp]
\centering
\caption{Example acronyms, misspelling words and their correct form in the acronym and spelling correction dictionary.}
\label{tab6}
\begin{tabular}{lll}
\hline
\begin{tabular}[c]{@{}c@{}}{\bf Acronyms, abbreviates} \\ {\bf and misspelling words}\end{tabular} & \begin{tabular}[c]{@{}c@{}}{\bf Vietnamese correct} \\ {\bf form}\end{tabular} & {\bf English} \\ \hline
cóa, coá                       & có       & have         \\ 
ngta,  nta                     & người ta & other people \\ 
cf, coffee, cafe, caphe, coffe & cà phê   & coffee       \\ 
cũnh, cungc, cungz             & cũng     & also         \\ 
pk, bit, bjt                   & biết     & know         \\ \hline
\end{tabular}
\end{table}

After correcting spelling mistakes, we then perform statistical analysis on UIT-VSMEC. Table \ref{tab_statistical} shows the results of the analysis.

\begin{table}[htbp]
\centering
\caption{Examples of word statistics.}
\label{tab_statistical}
\resizebox{\columnwidth}{!}{%
\begin{tabular}{lrrrrrrrr}
\hline
\textbf{Word} &
  \multicolumn{1}{c}{\textbf{Other}} &
  \multicolumn{1}{c}{\textbf{Disgust}} &
  \multicolumn{1}{c}{\textbf{Enjoyment}} &
  \multicolumn{1}{c}{\textbf{Anger}} &
  \multicolumn{1}{c}{\textbf{Surprise}} &
  \multicolumn{1}{c}{\textbf{Sadness}} &
  \multicolumn{1}{c}{\textbf{Fear}} &
  \multicolumn{1}{c}{\textbf{Total}} \\ \hline
\begin{tabular}[c]{@{}c@{}}cười\\ (laugh)\end{tabular}   & 248 & 279 & 1,173 & 95  & 77 & 114 & 99 & 2,085 \\ \hline
\begin{tabular}[c]{@{}c@{}}không\\ (no/not)\end{tabular} & 317 & 344 & 265  & 132 & 40 & 299 & 90 & 1,487 \\ \hline
\begin{tabular}[c]{@{}c@{}}vì\\ (because)\end{tabular}   & 29  & 46  & 42   & 16  & 0  & 51  & 9  & 193  \\ \hline
\begin{tabular}[c]{@{}c@{}}rất \\ (very)\end{tabular}    & 17  & 18  & 69   & 2   & 1  & 19  & 10 & 136  \\ \hline
\begin{tabular}[c]{@{}c@{}}tuổi\\ (age)\end{tabular}     & 18  & 14  & 30   & 7   & 1  & 19  & 7  & 96   \\ \hline
\end{tabular}%
}
\end{table}

As for technique 7, we select a list of words to test. These words satisfy three conditions: appear more than 14 times in total; appear at least five times in each emotion label; are not noun/noun phrases, verbs, adjectives or adverbs. The test list consists of more than 300 words. We believe training a classifier with full vocabulary will keep a text’s context. Then, filtering some stop words and particular words in the validation and test set will ignore the effect of these words. If we also remove these words from the train set, then our classifier will not understand a comment’s context. We evaluate the impact of the removal of these words on the validation set using an emotion classifier built on the cleaned train set.

The first test case is to filter out each word from the test list separately and check for the difference in F1-score. If removing a word leads to a significant decrease in F1-score, then it will be removed from the test list. On the other hand, if the removal of a word gives a higher F1-score, then these words will be moved to a final filter list. The test list now should consist of words that do not impact the F1-score on removal. The second and latter test cases will first filter out words that in the final filter list, then filter each word in the test list and repeat the process. If a test case ends with no new word in the final filter list, then the next test case will filter one additional word from the test list, but if this happens three times on a row (calculation limitation), technique 7 will end.

Technique 7 aims to find out a list of words that should be removed; these words are a combination of some stop words and some particular words that should be removed in emotion recognition task.

\subsection{Emotion Classifier}\label{MLR}

In this paper, we use Count Vectorizer and TF-IDF Vectorizer to vectorize input data and then use Multinomial Logistic Regression (MLR) \cite{ellen2011all} to train 22 emotion classifiers for Vietnamese microtext. Besides, Logistic Regression also achieved the best score on the VLSP shared task - Hate Speech Detection on Vietnamese social media texts \cite{vu2020hsd}. These classifiers are trained by the data that applied different pre-processing techniques. We compare the performance of each classifier to indicate suitable pre-processing methods for our task.
\subsection{Key-clause Extraction}\label{key-clause}

One common point of a text or a speech is that they will have at least one concluding sentence, in which it helps to identify key information. Furthermore, if the concluding sentence contains a coordinating conjunction, such as “but”, then this sentence can be separated into two or more clauses /phrases. For example, in the sentence “I cannot cook very well, but I make quite good fried egg”, both “I cannot cook very well” and “I make good fried egg” are main/independent clauses (they are of equal importance and could each exist as a separate sentence).

We propose an idea that exploits this language’s characteristic to filter out unimportant clauses in a comment. The idea is to rank (or give weight to) all sentences within a comment then rank the clauses within a sentence and remove clause of a specific rank. A sentence describing a feeling should have a higher rank than a sentence describing a place. The ideal situation is every sentence in a comment can be split into clauses, we then apply an emotion classifier on clause-level and then ensemble result based on clause’s rank to get a final prediction. If a comment itself is a sentence and cannot be split into clauses, then this method cannot be applied.

We have not achieved the final goal of this idea; instead, we come up with a method for extracting key-clause in a sentence. First, we separate each comment into clauses/phrases by using punctuation marks (comma, dot and semi-colon) and coordinating conjunction. If a separated clause contains less than four words, then this clause will be concatenated with its preceding or succeeding clause. 

Then, we apply the previously built emotion classifier on each clause of the same comment. Using the emotion prediction on each clause, we create a list of important words; these words appear many times in a correctly classified clause. Example of these words is “nhưng” (but), “tuy nhiên” (however), “đúng đắn nhất” (most righteous). We use these important words to indicate if a clause is the main clause or not. Table \ref{tab7} shows three examples of phrases extraction.

\end{otherlanguage*}

\begin{table}[htbp]
\centering
\caption{Phrases extraction for each comment.}
\label{tab7}
\centerline{\includegraphics[scale=0.67]{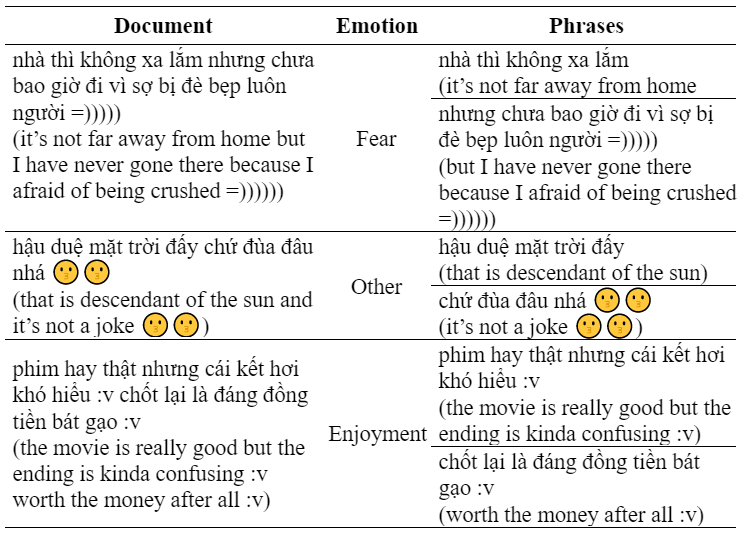}}
\end{table}

\section{Experiment Results}\label{Experiments}

\subsection{Experiments with pre-processing techniques}\label{compare-preprocessing}

We conduct several experiments, presented in Table \ref{tab8} and Table \ref{tab9}, to compare the performance of the pre-processing techniques described in Section \ref{pre-processing}. We use the default setting for Count Vectorizer, TF-IDF Vectorizer, and Multinomial Logistic Regression classifier in these experiments. We also use data in the train set, and the validation set to compare each pre-processing technique’s performance and to tune hyper-parameters. Finally, we use the test set to report final, unbiased classifier performance results in the paper. Our purpose is to indicate which vectorizer, pre-processing techniques, and parameter setting best fit our training purpose. Table \ref{tab8} shows the impact of pre-processing on model performance.

\begin{table}[htbp]
\centering
\caption{Impact of pre-processing techniques on the validation set of UIT-VSMEC.}
\label{tab8}
\begin{tabular}{clrr}
\hline
{\bf Vectorizer}            & {\bf Pre-processing} & {\bf Accuracy}         & {\bf F1-score}         \\ \hline
\multirow{9}{*}{Count Vectorizer}   & Original                  & 56.41\%          & 56.07\%          \\ 
  & 1                         & 54.81\%          & 54.85\%          \\ 
  & 1 + 2                     & 54.66\%          & 54.16\%          \\ 
  & 1 + 3                     & 55.98\%          & 55.55\%          \\ 
  & 1+ 3 + 5                  & 58.75\%          & 58.17\%          \\ 
  & \textbf{1 + 3 + 5 +6}     & \textbf{58.60\%} & \textbf{58.07\%} \\ 
  & 1 + 4                     & 57.00\%          & 56.58\%          \\ 
  & 1 + 4 +5                  & 58.45\%          & 57.87\%          \\
  & \textbf{1 + 4 + 5 + 6}    & \textbf{58.75\%} & \textbf{58.25\%} \\ \hline
\multirow{9}{*}{TF-IDF Vectorizer} & Original                  & 55.69\%          & 53.39\%          \\ 
  & 1                         & 55.83\%          & 53.60\%          \\ 
  & 1 + 2                     & 55.39\%          & 53.15\%          \\  
  & 1 + 3                     & 57.29\%          & 55.16\%          \\ 
  & 1 + 3 + 5                 & 59.04\%          & 57.29\%          \\ 
  & \textbf{1 + 3 + 5 + 6}    & \textbf{59.33\%} & \textbf{57.99\%} \\ 
  & 1 + 4                     & 57.00\%          & 54.84\%          \\  
  & 1 + 4 + 5                 & 58.89\%          & 57.34\%          \\  
  & \textbf{1 + 4 + 5 + 6}    & \textbf{58.60\%} & \textbf{56.94\%} \\ \hline
\end{tabular}
\end{table}

As shown in Table \ref{tab8}, technique 2 - removing emojis performs worse than technique 3 - keeping emojis and emoticons and then transforming them into the word form (more than 1\% different). The reason for this is because Facebook users tend to use emojis, emoticons in their comment, message, and even react to a post with emojis and reactions. The accuracy and F1-score of the model using pre-processing techniques 1, 3, 5, and 1, 4, 5 surpass the original model for each vectorizer type, proving technique 5 to be effective. 

Adding technique 6 leads to a slight decrease in accuracy and F1-score in the classifier model that uses Count Vectorizer and pre-processing techniques 1, 3, 5, and classifier models that use TF-IDF Vectorizer and pre-processing technique 1, 4, 5. In contrast, adding technique 6 slightly increases the accuracy and F1-score of the Count Vectorizer model using technique 1, 4, 5, and in the TF-IDF Vectorizer model using technique 1, 3, 5. Although the model that uses Count Vectorizer and pre-processing techniques 1, 4, 5, 6 achieves the best F1-score, the model that uses TF-IDF Vectorizer and pre-processing techniques 1, 3, 5, 6 achieve the best accuracy.

Therefore, we perform hyper-parameter tuning to find out which combination of vectorizer, pre-processing techniques and parameter setting that lead to the best performing model.

\subsection{Experiments with hyper-parameter fine-tuning}\label{hyper-parameter-tuning}
\begin{otherlanguage*}{vietnamese}
As a result of the previous experiment, we set the parameter C = 4.5 and class\_weight = “balanced” for the MLR classifier. In this experiment, we perform hyper-parameter tuning with the n\_features (ranging from 3,000 to 50,000) and ngram\_range (ranging from 1-1, 1-2 and 1-3 gram) parameter for the Count Vectorizer and TF-IDF Vectorizer. There are several reasons for these ranging. Firstly, UIT-VSMEC has a small vocabulary (4,097 words after pre-processing). Secondly, we do not perform word segmentation on the dataset. Lastly, Vietnamese words comprise of single word and compound word. Sometimes a single word does not have any meaning unless they go with another single word to form a compound word (e.g. “lác” and “đác” form the word “lác đác” - “little”). In contrast, some compound words are formed by words that have the meaning of their own, for example, “tiền công” (wage) is formed by “tiền” (money) and “công” (effort). We even use ngram\_range from 1-3 gram because the use of compound words, for example, “phim” (movie) and “Ấn Độ” (India) form the noun phrase “phim Ấn Độ” (Indian movie). The best results are presented in table \ref{tab9} along with its parameter setting.

\begin{table}[htbp]
\centering
\caption{Impact of hyper-parameter tuning on the development set of UIT-VSMEC.}
\label{tab9}
\resizebox{\columnwidth}{!}{%
\begin{tabular}{cccrr}
\hline
\begin{tabular}[c]{@{}c@{}}{\bf Feature} \\ {\bf Engineering}\end{tabular} &
  {\bf Pre-processing} &
  \begin{tabular}[c]{@{}c@{}}{\bf Vectorizer} \\ {\bf parameter}\end{tabular} &
  {\bf Accuracy} &
  {\bf F1-score} \\ \hline
\multirow{2}{*}{\begin{tabular}[c]{@{}c@{}}Count \\ Vectorizer\end{tabular}} &
  1 + 3 + 5 + 6 &
  \begin{tabular}[c]{@{}c@{}}n\_features\\ = 38,000\\ ngram\_range\\ = (1, 2)\end{tabular} &
  60.35\% &
  60.33\% \\ \cline{2-5} 
 &
  1 + 4  + 5 + 6 &
  \begin{tabular}[c]{@{}c@{}}n\_features\\ = 40,000\\ ngram\_range\\ = (1, 2)\end{tabular} &
  60.06\% &
  60.06\% \\ \hline
\multirow{2}{*}{\begin{tabular}[c]{@{}c@{}}TF-IDF \\ Vectorizer\end{tabular}} &
  \textbf{1 + 3 + 5 + 6} &
  \textbf{\begin{tabular}[c]{@{}c@{}}n\_features\\ = 25,000\\ ngram\_range\\ = (1, 3)\end{tabular}} &
  \textbf{61.22\%} &
  \textbf{61.04\%} \\ \cline{2-5} 
 &
  1 + 4 + 5 + 6 &
  \begin{tabular}[c]{@{}c@{}}n\_features\\ = 26,000\\ ngram\_range\\ = (1, 3)\end{tabular} &
  60.35\% &
  60.17\% \\ \hline
\end{tabular}%
}
\end{table}
\end{otherlanguage*}

Table \ref{tab9} shows that keeping repeated emojis performs slightly better than only keep the unique ones. The best accuracy and F1-score is the emotion classifier that uses TF-IDF Vectorizer. Therefore, we will use the TF-IDF Vectorizer with pre-processing technique 1, 3, 5, 6, and the parameters for the TF-IDF vectorizer are n\_features = 25,000, ngram\_range = (1,3), and the parameters for the classifier are C = 4.5 and class\_weight = “balanced” to build our classifier model. Thus, we will use this same set-up to train our baseline model on raw data.

\subsection{Experiments with pre-processing technique 7 and key-clause extraction method}\label{experiment-pp7-key-clause}
We use the classifier model to evaluate pre-processing technique 7, which removes stop words and particular words, and the performance of our key-clause extraction method. The results are shown in Table \ref{tab10}.

\begin{table}[htbp]
\centering
\caption{Experimental results of all model on the test set of UIT-VSMEC.}
\label{tab10}
\begin{tabular}{lrr}
\hline
\multicolumn{1}{c}{\textbf{Model}}      & \textbf{Accuracy} & \textbf{F1-score} \\ \hline
MLR (baseline)                                  & 55.41\%           & 55.57\%           \\ \hline
CNN + word2Vec \cite{ho2019emotion}                            & 59.74\%           & 59.74\%           \\\hline 
MLR + pre-processing techniques (1, 3, 5, 6) & 62.91\%           & 62.95\%           \\ \hline
\begin{tabular}[c]{@{}l@{}}MLR + pre-processing techniques (1, 3, 5, 6) \\ + remove stop words (7)\end{tabular}   & 64.07\%          & 64.07\%          \\ \hline
\begin{tabular}[c]{@{}l@{}}MLR+ pre-processing techniques (1, 3, 5, 6, 7) \\ + key-clause extraction\end{tabular} & \textbf{64.36\%} & \textbf{64.40\%} \\ \hline
\end{tabular}
\end{table}
Our first observation is that the model that uses pre-processing techniques 1, 3, 5, 6 outperforms the baseline by a large margin. This observation shows the power and effectiveness of data cleaning when working with Vietnamese microtext. There are two reasons for this. First, the pre-processing techniques reduce the vocabulary size by normalizing text data; it means a portion of noise data filter out. Second, we also take advantage of the emojis and emoticons by transforming them into their word form. We consider these are social media data's characteristics.

The next observation is the impact of the removal of stop words and particular words. Removing stop words yields a higher F1-score (1.12\%). This result is reasonable because these words are considered noise data if we want to detect a comment’s emotion.

Finally, the implementation of the key-clause extraction method affects the performance of the emotion classifier. The explanation for this is that the clauses extracted from a comment by this method are most likely to carry the emotion. Another explanation is because our proposed method is not completed; it only affect 23 out of 693 comments in the test set.

Our best model using pre-processing techniques 1, 3, 5, 6, 7, and key-clause extraction methods achieve 64.40\% in the F1 score, which improves 8.83\% compared with the baseline model. Our emotion classifier model outperform the CNN model built by \cite{ho2019emotion} with an improvement of 4.66\% in F1-score. The reason behind this is because of the implementation of pre-processing methods on UIT-VSMEC. The results also confirm the effectiveness of using the key-clause extraction method.

\subsection{Error Analysis}\label{error-analysis}

Table \ref{tab11} shows some examples that our model fails to detect the correct emotion label and our explanations. Typical cases include: 1) our model does not understand the context; 2) our model does not understand words that have never appeared before; 3) our model does not filter emoticons and emojis unrelated to the comment; 4) our model recognizes a wrong clause as key-clause.

\begin{otherlanguage*}{vietnamese}
\begin{table}[htbp]
\centering
\caption{Some error cases.}
\label{tab11}
\begin{tabular}{lcll}
\hline
\textbf{Comments} &
  \textbf{Emotion} &
  \multicolumn{1}{c}{\textbf{Predictions}} &
  \textbf{Explanations} \\ \hline
\begin{tabular}[c]{@{}l@{}}người ta có bạn \\ bè nhìn vui thật \\ (they look so \\ happily having \\ friends)\end{tabular} &
  Sadness &
  Enjoyment &
  \begin{tabular}[c]{@{}l@{}}Our model cannot \\ understand the \\ context of this \\ comment.\end{tabular} \\ \hline
\begin{tabular}[c]{@{}l@{}}nghe ngọt thế :((\\ (sounds so sweet \\ :(( )\end{tabular} &
  Enjoyment &
  Sadness &
  \begin{tabular}[c]{@{}l@{}}The ":(("emoticon \\ have the strongest \\ influence in this \\ comment.\end{tabular} \\ \hline
\begin{tabular}[c]{@{}l@{}}kinh khủng thật :((\\ (it's terrilble:(( )\end{tabular} &
  Fear &
  Sadness &
  \begin{tabular}[c]{@{}l@{}}The ":(("emoticon \\ have a stronger \\ influence than the \\ word "kinh khủng" \\ (terrible).\end{tabular} \\ \hline
\begin{tabular}[c]{@{}l@{}}tao đéo ngờ được \\ đây là trường tao \\ :))))\\ (I did not expect \\ this was my school\\ :))))))\end{tabular} &
  Surprise &
  Enjoyment &
  \begin{tabular}[c]{@{}l@{}}The term "đéo ngờ \\ được" (did not expect) \\ has not appeared in \\ the train data so it \\ cannot be recognized \\ to express \\ surpriseness.\end{tabular} \\ \hline
\end{tabular}
\end{table}
\end{otherlanguage*}
\section{Conclusion and Future work}\label{conclusion}
In this study, based on Vietnamese social media characteristics, we explore the effect of our pre-processing techniques and our key-clause extraction technique combined with the machine learning model toward the textual emotion recognition task in Vietnamese social media text. We also reach the best overall weighted F1-score of 64.40\% on the pre-processed UIT-VSMEC corpus with Multinomial Logistic Regression using the TF-IDF Vectorizer and key-clause extraction method.

There are some limitations to our works that we will discuss in the following. First, UIT-VSMEC is not generalized enough. The dataset needs to have more cases about anger, fear, and surprise labels, and the “Other” label needs to be more specific if we want to build a better system for real-life tasks. Second, due to language knowledge limitations, we cannot pre-process the data well enough. Finally, our work focuses only on recognizing six basic emotion \cite{ekman1999basic} labels with the “other” label that is not included. 

As future work, we plan to improve the key-clause extraction method’s performance with semantic and syntax techniques. Besides, we also aim to build a larger, manually annotated dataset with fine-grained emotions following the study \cite{demszky-etal-2020-goemotions}. We will apply transfer models such as BERT \cite{devlin2019bert} and its variations to improve performance of Vietnamese emotion recognition. 
\printbibliography

@article{ellen2011all,
  title={All about microtext},
  author={Ellen, Jeffrey},
  journal={A Working Definition and a Survey of Current Microtext Research within Artificial Intelligence and Natural Language Processing, USA: Sciterpress (Science and Technology Publications, Lda)},
  pages={329--336},
  year={2011}
}

@inproceedings{brilis2012mood,
  title={Mood classification using lyrics and audio: A case-study in greek music},
  author={Brilis, Spyros and Gkatzou, Evagelia and Koursoumis, Antonis and Talvis, Karolos and Kermanidis, Katia L and Karydis, Ioannis},
  booktitle={IFIP International Conference on Artificial Intelligence Applications and Innovations},
  pages={421--430},
  year={2012},
  organization={Springer}
}

@inproceedings{ho2012high,
  title={A high-order hidden Markov model for emotion detection from textual data},
  author={Ho, Dung T and Cao, Tru H},
  booktitle={Pacific Rim Knowledge Acquisition Workshop},
  pages={94--105},
  year={2012},
  organization={Springer}
}

@article{ghazi2014prior,
  title={Prior and contextual emotion of words in sentential context},
  author={Ghazi, Diman and Inkpen, Diana and Szpakowicz, Stan},
  journal={Computer Speech \& Language},
  volume={28},
  number={1},
  pages={76--92},
  year={2014},
  publisher={Elsevier}
}

@article{da2014tweet,
  title={Tweet sentiment analysis with classifier ensembles},
  author={Da Silva, Nadia FF and Hruschka, Eduardo R and Hruschka Jr, Estevam R},
  journal={Decision Support Systems},
  volume={66},
  pages={170--179},
  year={2014},
  publisher={Elsevier}
}

@article{ho2019emotion,
  title={Emotion recognition for vietnamese social media text},
  author={Ho, Vong Anh and Nguyen, Duong Huynh-Cong and Nguyen, Danh Hoang and Pham, Linh Thi-Van and Nguyen, Duc-Vu and Van Nguyen, Kiet and Nguyen, Ngan Luu-Thuy},
  journal={Computational Linguistics - 16th International Conference of the Pacific Association for Computational Linguistics},
  year={2019}
}

@article{sailunaz2018emotion,
  title={Emotion detection from text and speech: a survey},
  author={Sailunaz, Kashfia and Dhaliwal, Manmeet and Rokne, Jon and Alhajj, Reda},
  journal={Social Network Analysis and Mining},
  volume={8},
  number={1},
  pages={28},
  year={2018},
  publisher={Springer}
}

@article{calvo2010affect,
  title={Affect detection: An interdisciplinary review of models, methods, and their applications},
  author={Calvo, Rafael A and D'Mello, Sidney},
  journal={IEEE Transactions on affective computing},
  volume={1},
  number={1},
  pages={18--37},
  year={2010},
  publisher={IEEE}
}

@article{ekman1999basic,
  title={Basic emotions},
  author={Ekman, Paul},
  journal={Handbook of cognition and emotion},
  volume={98},
  number={45-60},
  pages={16},
  year={1999}
}

@article{bohning1992multinomial,
  title={Multinomial logistic regression algorithm},
  author={B{\"o}hning, Dankmar},
  journal={Annals of the institute of Statistical Mathematics},
  volume={44},
  number={1},
  pages={197--200},
  year={1992},
  publisher={Springer}
}

@inproceedings{satapathy2017phonetic,
  title={Phonetic-based microtext normalization for twitter sentiment analysis},
  author={Satapathy, Ranjan and Guerreiro, Claudia and Chaturvedi, Iti and Cambria, Erik},
  booktitle={2017 IEEE International Conference on Data Mining Workshops (ICDMW)},
  pages={407--413},
  year={2017},
  organization={IEEE}
}

@article{vu2020hsd,
  title={HSD shared task in VLSP campaign 2019: Hate speech detection for social good},
  author={Vu, Xuan-Son and Vu, Thanh and Tran, Mai-Vu and Le-Cong, Thanh and Nguyen, Huyen},
  journal={arXiv preprint arXiv:2007.06493},
  year={2020}
}

@inproceedings{demszky-etal-2020-goemotions,
    title = "{G}o{E}motions: A Dataset of Fine-Grained Emotions",
    author = "Demszky, Dorottya  and
      Movshovitz-Attias, Dana  and
      Ko, Jeongwoo  and
      Cowen, Alan  and
      Nemade, Gaurav  and
      Ravi, Sujith",
    booktitle = "Proceedings of the 58th Annual Meeting of the Association for Computational Linguistics",
    month = jul,
    year = "2020",
    address = "Online",
    publisher = "Association for Computational Linguistics",
    url = "https://www.aclweb.org/anthology/2020.acl-main.372",
    doi = "10.18653/v1/2020.acl-main.372",
    pages = "4040--4054",
}

@inproceedings{devlin2019bert,
  title={BERT: Pre-training of Deep Bidirectional Transformers for Language Understanding},
  author={Devlin, Jacob and Chang, Ming-Wei and Lee, Kenton and Toutanova, Kristina},
  booktitle={NAACL-HLT (1)},
  year={2019}
}
\end{document}